\pdfoutput=1

\documentclass[11pt]{article}

\usepackage {ACL2023}

\usepackage{times}
\usepackage{latexsym}

\usepackage[T1]{fontenc}

\usepackage[utf8]{inputenc}

\usepackage{microtype}

\usepackage{inconsolata}
\usepackage{graphicx}
\usepackage{amsmath}
\usepackage{adjustbox}
\usepackage{multirow}
\usepackage{svg}
\usepackage[normalem]{ulem}
\useunder{\uline}{\ul}{}
\usepackage{makecell}
\usepackage{tablefootnote}

\usepackage{xcolor}

\definecolor{colorhigh}{HTML}{9900FF}
\definecolor{colorlow}{HTML}{660000}

\title{Leveraging Domain Adaptation and Data Augmentation to Improve Qur'anic IR in English and Arabic}

\author{Vera Pavlova \\
  rttl.ai \\ Dubai, UAE \\
  \texttt{v@rttl.ai} \\}

\begin{document}
\maketitle
\begin{abstract}
In this work, we approach the problem of Qur'anic information retrieval (IR) in Arabic and English. Using the latest state-of-the-art methods in neural IR, we research what helps to tackle this task more efficiently.
Training retrieval models requires a lot of data, which is difficult to obtain for training in-domain. Therefore, we commence with training on a large amount of general domain data and then continue training on in-domain data. To handle the lack of in-domain data, we employed a data augmentation technique, which considerably improved results in MRR@10 and NDCG@5 metrics, setting the state-of-the-art in Qur'anic IR for both English and Arabic. The absence of an Islamic corpus and domain-specific model for IR task in English motivated us to address this lack of resources and take preliminary steps of the Islamic corpus compilation and domain-specific language model (LM) pre-training, which helped to improve the performance of the retrieval models that use the domain-specific LM as the shared backbone. We examined several language models (LMs) in Arabic to select one that efficiently deals with the Qur'anic IR task. Besides transferring successful experiments from English to Arabic, we conducted additional experiments with retrieval task in Arabic to amortize the scarcity of general domain datasets used to train the retrieval models. Handling Qur'anic IR task combining English and Arabic allowed us to enhance the comparison and share valuable insights across models and languages.

\end{abstract}

\begin{figure}[h]
\centering
\includegraphics[width=0.9\columnwidth]{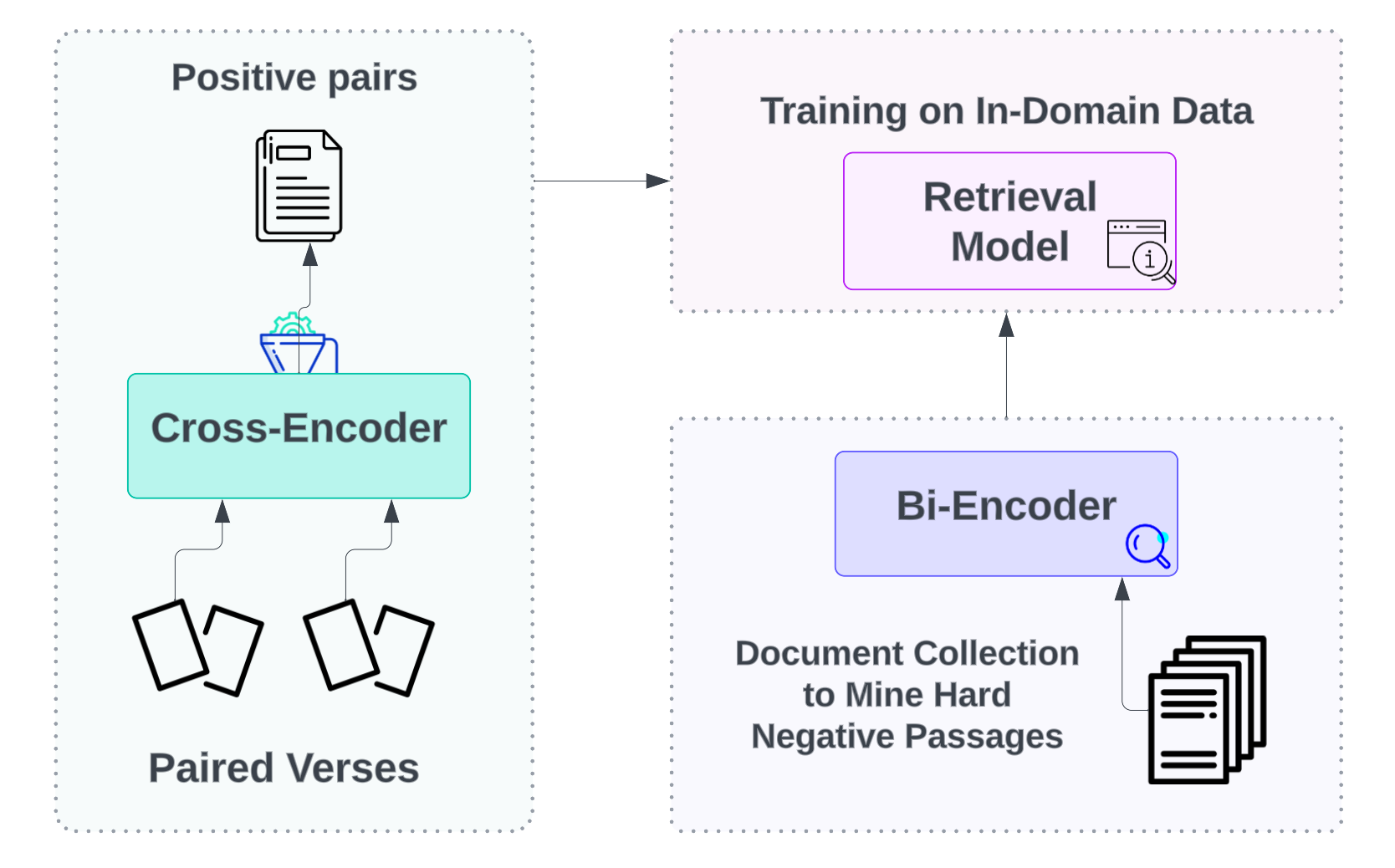}
\caption{Data augmentation technique that leverages correlation of Qur'anic verses for training retrieval models in-domain.}
\label{fig:figure 1}
\end{figure}

\section{Introduction}

Recent advances in Natural Language Processing (NLP) have helped to improve search relevance and retrieval quality. Nevertheless, deep-learning techniques, specifically transformer-based approaches \citep{10_vaswani_2017}, are hardly employed in Quran'ic NLP \cite{bashir2023arabic}. In this work, we will utilize the latest state-of-the-art neural retrieval models to compare what works best for solving the IR task in the Islamic domain. Moreover, we proposed a data-augmentation technique to generate data for in-domain training appropriate for the IR task involving the Holy Qur'an (see Figure ~\ref{fig:figure 1}). 

We experimented with Arabic and English languages. Arabic, more precisely one of its variants, Classical Arabic (CA), is the language of the Holy Qur'an and is an integral component in tackling retrieval task using sacred scripture \cite{bashir2023arabic}. English is another popular language used for search in various domains, including the Islamic domain. Addressing the problem using Arabic and English allows for comparing the solutions and sharing insights across languages. English is a high-resource language with a great choice of corpora and pre-trained LMs for diverse domains. At the same time, depending on the domain, Arabic can be considered a low- or medium-resource language (\citealp{xue-etal-2021-mt5}; \citealp{abboud-etal-2022-cross}).
However, Arabic is in more favorable conditions than English in the Islamic domain; in the case of the Arabic language, there are Islamic corpora like OpenITI \cite{romanov2019openiti} and domain-specific LMs (\citealp{MALHAS2022103068}; \citealp{inoue-etal-2021-interplay}). From this perspective, addressing Qur'anic IR in English is more challenging as it requires a number of additional preparations, like preparing an Islamic corpus and pre-training domain-specific LM. This state of affairs with the English language in the Islamic domain necessitates addressing it alongside the Arabic language. Simultaneously, another advantage of handling the problem in English is the abundance of datasets to train for a general domain. Training on general domain data can be a required step to prepare a retrieval model that needs a substantial amount of data for training, where in-domain data is scarce. Experimenting with Qur'anic IR in English will allow us to learn what works best and apply these approaches to Arabic, where general domain data is insufficient.  

Our main contributions are:
\begin{itemize}
    \item We introduce an Islamic corpus and a new language model for the Islamic domain in English.
    \item We conduct comprehensive experiments with different retrieval models to see what works best for efficient retrieval from the Holy Qur'an in Arabic and English.
    \item We propose a data-augmentation technique that helped to improve the retrieval models' performance for both languages and set a new state-of-the-art in Qur'anic IR.
\end{itemize}
The rest of the work is organized as follows: we start with addressing the problem of Qur'anic IR in English. We prepare the Islamic corpus and domain-specific LM (Section~\ref{sec:English LM}). Section ~\ref{sec:English IR} applies to both languages, English and Arabic, including metrics choice, datasets for training and testing, experimental details, and training procedure of the retrieval models. Section ~\ref{sec:Arabic IR} is dedicated to Qur'anic IR in Arabic. Apart from applying successful experiments that worked well with Qur'anic IR in English, we executed more methods of preparing retrieval models for Arabic language, including teacher-student distillation and employing machine translation. Model comparison and Final analysis is performed in Section ~\ref{sec:Final}. The prior work done in the field is highlighted in Section ~\ref{sec:related work}.

\begin{figure}[h]
\centering
\includegraphics[width=0.9\columnwidth]{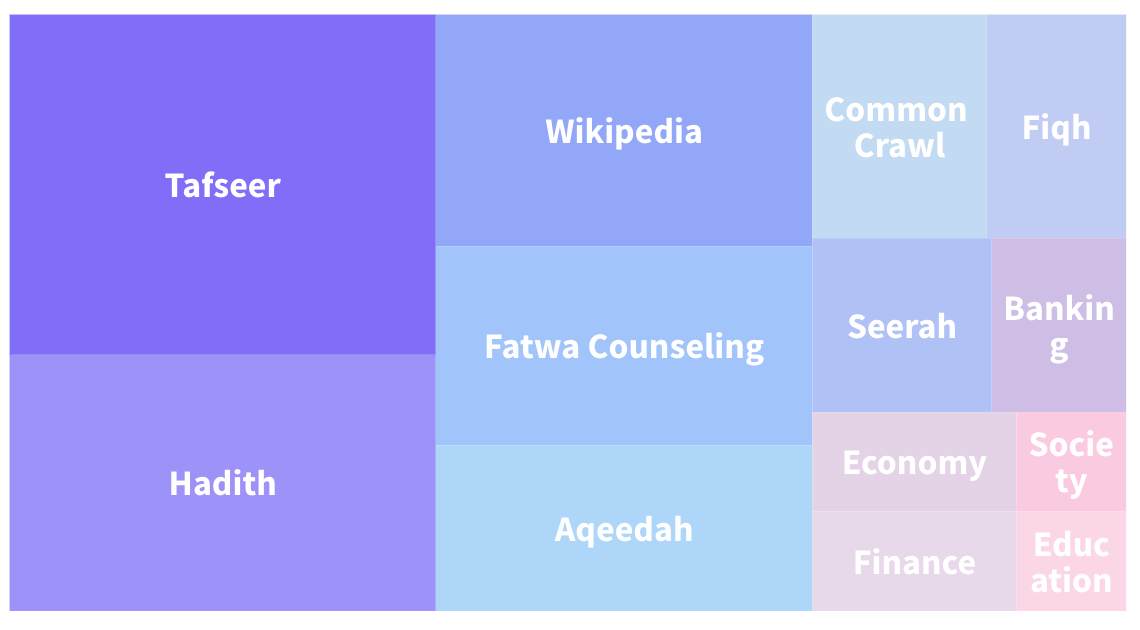}
\caption{Types of Islamic text that constitute Islamic Corpus.}
\label{fig:figure 2}
\end{figure}

\section{Domain-Specific Language Model as a Backbone of In-Domain IR}
\label{sec:English LM}

\subsection{Islamic Corpus in English}
\label{sec:Islamic Corpus}

Preparing an Islamic Corpus in English is challenging due to the insufficient amount of Islamic Text that is either translated from Arabic or other languages to English or initially written in English. We collect text available online of the following types (see Figure ~\ref{fig:figure 2}):

\textbf{Islamic literature.} These are Islamic books written by Islamic scholars about Tafseer (Qur'an exegesis), Hadith, Seerah, Fiqh (Islamic jurisprudence), and Aqeedah (Islamic creed) (approx. 28M words).

\textbf{Islamic journals.} Journals that are available online and focus on discussing modern issues of Islamic banking, Finance, Economy, and Islamic Education (approx. 5.5 M words).

\textbf{Fatwa counseling.} Fatwas that are available online from Fatwa centers (approx. 4.8M words)

\textbf{Wikipedia.} Articles related to Islam from the Wikipedia Islam portal (approx. 5.6M words).

\textbf{Common Crawl.} We search for keywords and collect files from Common Crawl on Islamic topics. We perform additional filtering and preprocessing of these articles (approx. 2.5 M words).

The total amount of words in the corpus is around 47M words. 

\subsection{Adaptation of General Domain Language Model for Islamic Domain }
\label{sec:Domain Adaptation}

Pre-training starting from the existing checkpoint of the model pre-trained for the general domain helps reduce pre-training time (\citealp{gururangan-etal-2020-dont}; \citealp{bommasani_2022_opportunities}; \citealp{guo2022domain}). To account for the small size of the pre-training corpus and perform domain adaptation effectively, we continue pre-training the BERT model on the Islamic corpus.
To address the issue of the absence of domain-specific vocabulary during continued pre-training, we trained the WordPiece tokenizer \citep{song-etal-2021-fast} on the Islamic corpus. We find the intersection between Islamic vocabulary and bert-base-uncased, and for the tokens inside this intersection, we assign the weights from bert-base-uncased.\footnote{\label{bert}
{\url{https://huggingface.co/bert-base-uncased}}} For the tokens outside of the intersection (Islamic tokens), we perform contextualized weight distillation following \citep{pavlova-makhlouf-2023-bioptimus}. \footnote{\url{https://github.com/rttl-ai/BIOptimus}}
\begin{itemize}
    \item In the first step, we find tokens of interest and extract corresponding sentences from the Islamic Corpus. We sample from one to twenty sentences \citep{bommasani-etal-2020-interpreting}.
    \item In the second step, we tokenize these sentences using a bert-base-uncased tokenizer. In that case, Islamic tokens are broken into subtokens because they are absent from bert-base-uncased vocabulary. We average the contextualized weights of the corresponding subtokens that the BERT model produces ($t_{distilled}$) and then compute aggregated representation across sentences ($t_{aggregated}$) for a corresponding token of interest from Islamic vocabulary:
    \begin{equation}
    \begin{split}
        t_{distilled} = f(t_1,...,t_k) \\ f \in \{mean\}\\
    \end{split}
    \end{equation}
    Where \emph{k} is the number of the subtokens that make up the token of interest.
    \begin{equation}
    \begin{split}
        t_{aggregated} = g(t_{distilled},...,t_m) \\ g \in \{mean\}\\
    \end{split}
    \end{equation} 
    And \emph{m} is the number of sentences involved in aggregated representation. 
\end{itemize}
In order to avoid overinflating the vocabulary with new tokens, which would require longer pre-training and be prohibitive in the case of a small corpus, we analyze the frequency of each token in our corpus. Tokens with a count below threshold are filtered out, resulting in 3992 new domain-specific tokens. Moreover, we remove [unused] tokens from the bert-base-uncased vocabulary and add Islamic tokens, resulting in 33511 total tokens in the BPIT model's final vocabulary (BPIT is the abbreviation for BERT Pre-trained on Islamic Text).

\subsection{Pre-training Set-up}

In order to accommodate the limited size of the pre-training corpus, we schedule two-stage pre-training akin to phases of Curriculum learning (\citealp{bengio_2009}; \citealp{soviany2022curriculum}). In the first stage, we start with an easier task of predicting masked tokens/subtokens, with a masking rate of 0.15 and using the "80-10-10" corruption rule (\citealp{devlin-etal-2019-bert}; \citealp{wettig-etal-2023-mask}). 
In the second stage, we increase the complexity of the prediction task by switching to predicting the whole words with the same masking rate and using the corruption rule.
It is harder for a language model to predict whole words than to predict tokens or subtokens that might make up the word and give the LM more clues and make the prediction task less challenging (\citealp{9599397}; \citealp{dai-etal-2022-whole}; \citealp{10.1145/3458754}). This pre-training approach introduces the LM to a broader scope of language experience and helps to gain more diversified knowledge of textual input \citep{mitchell1997machine}, which is crucial in the case of a small corpus that we use. Pre-training hyperparameters can be found in Appendix ~\ref{sec:appendix}.

\section{Preparing Neural IR Model to Retrieve from the Holy Qur’an}
\label{sec:English IR}

\subsection{Dataset for Testing Retrieval Models}
\label{sec:test dataset}

To test our models, we converted the QRCD (Qur'anic Reading Comprehension Dataset) \citep{MALHAS2022103068} to the IR dataset. We use both train and development split as test data. We do not include no-answer questions \citep{malhas2020ayatec}, which results in 169 queries in total for testing. Queries are accompanied by the corresponding verses from the Holy Qur'an. Each Qur'anic verse is treated as the basic retrieval unit because it presents a more challenging task (see Section ~\ref{sec:Final}) and has higher utilization factors. The original dataset is in Arabic and was constructed and annotated by experts in Islamic studies. For our purposes of testing IR systems, we translate queries to English and verify the validity and accuracy of the translation with Islamic scholars. We use the Saheeh International translation of the Holy Qur’an to express specific Qur’anic terms used in query formulation. To retrieve answers, we use the same Sahih International translation as a retrieval collection.\footnote{\url{https://tanzil.net/trans/}} 

\subsection{Metrics}
\label{sec:metrics}

Due to the complexity of the language of the Holy Qur’an and the fact that some meanings can be expressed indirectly, the retrieval task using the Holy Qur’an is quite difficult. Therefore, using several metrics to estimate the retrieval model's effectiveness from a different perspective makes sense.
We use the MRR@10 (Mean Reciprocal Rate), the official evaluation metric of the MS MARCO dataset \citep{bajaj2018ms} that we extensively use to fine-tune our retrieval models. Furthermore, we add NDCG@5 (Normalized Discounted Cumulative Gain) and Recall@100, used in the BEIR benchmark \citep{thakur-2021-BEIR}. This combination of metrics lets us estimate our models with a decision support metric such as Recall, binary rank-aware metrics such as MRR, and metric with a graded relevance judgment such as NDCG \citep{wang2013theoretical}. For evaluation, we use the BEIR framework \footnote{\url{ https://github.com/beir-cellar/beir/tree/main}} that utilizes the Python interface of the TREC evaluation tool \citep{Van_Gysel_2018}.

\subsection{Baselines}
\label{sec:baselines}

BM25 is a commonly used baseline to compare retrieval systems. It is a sparse lexical retrieval method based on token-matching and uses TF-IDF weights. Though the lexical approaches suffer from the lexical gap \citep{Berger2000BridgingTL} due to the constraints of retrieving the documents containing exact keywords presented in a query, BM25 remains a strong baseline \citep{kamalloo2023resources}.
We also include a dense neural retrieval model,  trained using a sentence-transformers framework \citep{reimers-gurevych-2019-sentence} referred to as SBERT- GD (general domain), a late-interaction model ColBERT \citep{khattab2020colbert} (ColBERT-GD), and  Cross-Encoder-GD. All models were trained on the MS MARCO dataset from the bert-base-uncased checkpoint. This approach allows us to evaluate their performance in a zero-shot setting for the Islamic domain and compare them with the retrieval models trained using the domain-specific BPIT model. More details on how SBERT-GD,  ColBERT-GD, and Cross-encoder-GD were trained are presented in Section ~\ref{sec:BPIT}; hyperparameters details are listed in Appendix~\ref{sec:appendix}.

\subsection{Training a Domain-specific Model on General Domain Data}
\label{sec:BPIT}

To prepare the domain-specific model for the IR task, we prepare and compare three approaches.

\textbf{SBERT-BPIT}. We use the sentence-transformers framework, which employs a Siamese network \citep{10.5555/2987189.2987282} that enables semantic similarity search. We train our BPIT model using the architecture above on the MS MARCO dataset that contains 533k training examples (more details on MS MARCO dataset are in Section ~\ref{sec:MT}), utilizing Multiple Negative Ranking Loss (MNRL) (\citealp{henderson2017efficient}; \citealp{ma-etal-2021-zero}; \citealp{oord2019representation}). MNRL is a cross-entropy loss that treats relevant pairs $\{x^{(i)}, y^{(i)}\}_{i=1}^M$ (where \emph{M} is batch size) as positive labels and other in-batch examples as negative, and formally defined as:
\begin{align*}
J_{\mathrm{MNRL}}&(\theta) = \\\frac{1}{M}\sum_{i=1}^M&\log\frac{\exp{\sigma(f_{\theta}(x^{(i)}), f_{\theta}(y^{(i)}))}}{\sum_{j=1}^M\exp{\sigma(f_{\theta}(x^{(i)}), f_{\theta}(y^{(j)}))}}
\end{align*}

where $\sigma$ is a similarity function, in our case it is a cosine similarity function; $f_{\theta}$ is the sentence encoder. 
We use multiple hard negatives; these are negative passages similar to the positive passage but not relevant to the query and mined using cross-encoder scores.\footnote{\url{https://huggingface.co/datasets/sentence-transformers/msmarco-hard-negatives}} 

\textbf{Cross-encoder-BPIT}. In the case of a cross-encoder, a pair of sentences are simultaneously fed into a transformer-like model, and attention is applied across all tokens to produce a similarity score \citep{humeau2020polyencoders}. This approach does not allow end-to-end information retrieval and endure extreme computational overhead. However, in many IR tasks, it performs superior to other methods and can be used for mining hard negatives, data augmentation (Section ~\ref{SEC:IN-domain}), and reranking. The model is trained with triples provided by MS MARCO starting from the BPIT checkpoint under a classification task, using Cross Entropy Loss.

\textbf{ColBERT-BPIT}. ColBERT computes embeddings independently for queries and documents and, at the same time, can also register more fine-grained interactions between tokens. Using the same mined hard negatives constructed for the MS MARCO dataset used to pre-train SBERT-BPIT, ColBERT-BPIT is trained starting from the BPIT checkpoint by optimizing the cross-entropy loss applied to the score of the query and the positive passage against in-batch negatives \citep{santhanam-etal-2022-colbertv2}.

All models with the prefix \textbf{BPIT} are counterparts of \textbf{GD} models; for a fair comparison, they are trained using the same dataset, objective function, and hyperparameters (see Appendix ~\ref{sec:appendix}) with the only difference that \textbf{BPIT} models initialized from the BPIT checkpoint and \textbf{GD} models initialized with the bert-base-uncased checkpoint.

\subsection{In-domain Training of the Domain-specific Model}
\label{SEC:IN-domain}

The performance of dense retrieval systems worsens when encountering a domain shift \citep{thakur-2021-BEIR}; therefore, there is a great benefit in training neural IR models on in-domain data. The lack of domain-specific data is often solved by augmenting training data: generating synthetic data \citep{tanaka2019data}, paraphrasing using synonyms \citep{wei-zou-2019-eda}, sampling and recombining new training pairs \citep{thakur-etal-2021-augmented}, round-trip translation (\citealp{yu2018qanet}; \citealp{xie2020unsupervised}) or involving denoising autoencoders \citep{wang-etal-2021-tsdae-using}. These techniques involve data distortion, which is suboptimal when dealing with religious and heritage datasets.  We propose a data generation technique for in-domain training advantageous for the retrieval task involving the text of the Holy Qur'an (see Figure ~\ref{fig:figure 1}).
Understanding the text of the Holy Qur’an is closely related to the meaning explained in the books of Tafseer written by Islamic Scholars. Tafseer Ibn Kathir, one of the established books of Qur'an exegesis, contains ample verse relations references. Putting this into use allows not only to perform data augmentation but also to intertwine more meaning to Qur’anic verses that need to be more explicit for a LM to learn directly from the text of the verse.

\textbf{Pairing}. Let  $C_{t}$ denote a collection of books of Tafseer by Ibn Kathir. We start with extracting and paring all verse relations mentioned in Tafseer Ibn Kathir. That gives us $V_{t}$ that contains distinct pairs $\{v_{q}, v_{p}\} \in V_{t}$ and $\vert V_{t}\vert\sim$ 11k pairs.

\textbf{Filtering}. Not all the pairs can be used for training the retrieval model because not all the verse relation pairs will be interpreted by the model as a signal of positive correlation due to meanings that are expressed indirectly. We use the cross-encoder model $M_{ce}$ that was trained on a general domain to score ayah pairs $s_{ce} = M_{C}(v_{q},v_{p})$. We filter out the pairs that were scored below the threshold, leaving us with $V_{f}$ that contains pairs with strong positive correlations $(q,p^{+})\in V_{f}$ and $\vert V_{f}\vert = 2352$ pairs.

\textbf{Sampling hard negatives}. To prepare negative passages, we use the text of the Tafseer Ibn Kathir without verses' quotations. The text is split into $M$ passages to form a collection $C^- = \{p^{-}_{1}, p^{-}_{2}, ...,p^{-}_{m}\}$ to sample negative passages. Sampling from the Holy Qur'an text is a less favorable approach. Due to the relatively small size of the Qur'anic corpus, mined negative passages may turn out to be false negatives \citep{qu-etal-2021-rocketqa}. At the same time, sampling from another corpus would create easy negatives that are not beneficial for training (\citealp{ren-etal-2021-rocketqav2}; \citealp{karpukhin-etal-2020-dense}; \citealp{xiong2021answering}), while the text of the Tafseer Ibn Kathir contains passages that are similar to the positive passages but not precisely relevant to $q$ and are good candidates to play a role of hard negatives. To choose hard negatives, we use a retrieval model trained with a Seamise network $M_{B}$ and retrieve negative passages $(p^{-}_{1}, ... ,p^{-}_{i})$ related to $\forall q \in V_{f}$. We score each pair $(q,p^{-})$ with the cross-encoder $s_{ce} = M_{C} (q, p^{-})$, and use these scores in the next stage of training.

\textbf{Continue training in-domain}. We combine the collection of verses from the Holy Quran $C^+$ and the collection of passages from Tafseer Ibn Kathir $C^-$ into one collection $C_{aug}$ for training, which together with selected positive pairs and mined hard negatives forms new augmented dataset $D_{IN}$ for in-domain training. We continue training SBERT-BPIT and ColBERT-BPIT on new in-domain data following the same procedure described for each model in Section ~\ref{sec:BPIT}. The models that come out as a result of this stage of training are SBERT-ID (Islamic Domain) and ColBERT-ID.

\subsection{Results and Discussion}
The performance of all models on the test dataset is collected in Table ~\ref{tab:tabel 1}. All the models steadily outperform the BM25 baseline on every metric. 
In the category of the \textbf{GD} and \textbf{BPIT} models, the best-performing model is \textbf{ColBERT} for all metrics. In contrast, in the category of \textbf{ID} models, \textbf{SBERT} shows the best results at \textbf{MRR@10}, with a considerable improvement in performance after in-domain training on the augmented dataset (increasing from \textbf{0.48} to \textbf{0.55}). 

Overall, all \textbf{ID} models demonstrate superior performance, proving that training in-domain using our data augmentation technique was beneficial. 
Moreover, another important observation is consistent progress for \textbf{SBERT} and \textbf{ColBERT} models when training using the domain-specific model (\textbf{BPIT}) coupled with training on in-domain data. 
We suppose that leveraging  domain adaptation of a LM that serves as a backbone for retrieval models and subsequent training of retrieval models on large general domain data before training on in-domain data is an effective approach. 

In Table ~\ref{tab:tabel 2}, we included a comparison and analysis of the performance of the retrieval models for different translations of the Holy Qur'an into English. We can see no significant degradation of the models’ performance. The formulation of the queries contains terms from Saheeh International translation (Section \ref{sec:test dataset}), which proves
that the models can maintain search relevancy with different semantics. With these results and insights, we switch to exploring how to tackle IR tasks for the Holy Quran in the Arabic Language.



\begin{table}[]
\begin{adjustbox}{width=\columnwidth,center}
\begin{tabular}{cccc}
\hline
\multicolumn{1}{l}{}                       & \textbf{\begin{tabular}[c]{@{}c@{}} Recall@100\end{tabular}} & \textbf{\begin{tabular}[c]{@{}c@{}} MRR @10 \end{tabular}} & \textbf{\begin{tabular}[c]{@{}c@{}} NDCG@5 \end{tabular}} \\ \hline
\multicolumn{1}{c}{\textbf{BM25}}     & 0.15                                                                                                                           & 0.27                                                                                                           & 0.15                                                                                                       \\
\multicolumn{1}{c}{\textbf{SBERT-GD}}  & 0.2                                                                                                                           & 0.43                                                                                                           & 0.23                                                                                                       \\

\multicolumn{1}{c}{\textbf{ColBERT-GD}} & 0.25                                                                                                                     & 0.43                                                                                                                 & 0.26                                                                                                       \\
\multicolumn{1}{c}{\textbf{Cross-encoder-GD}}        &  0.19                                                                                                                     & 0.37                                                                                                                & 0.22                                                                                                     \\
\multicolumn{1}{c}{\textbf{SBERT-BPIT}}       & 0.28                                                                                                                     & 0.48                                                                                                                 & 0.28                                                             
                                             \\
\multicolumn{1}{c}{\textbf{ColBERT-BPIT}}       & 0.32                                                                                                                     & 0.51                                                                                                                 & 0.32  
                                               \\
\multicolumn{1}{c}{\textbf{Cross-encoder-BPIT}}       & 0.17                                                                                                                     & 0.3                                                                                                                 & 0.16                                 
                                           \\
\multicolumn{1}{c}{\textbf{SBERT-ID}}       & 0.32                                                                                                                     & \textbf{0.55}                                                                                                                 & \textbf{0.33}                              
                                          \\

\multicolumn{1}{c}{\textbf{ColBERT-ID}}       &  \textbf{0.33}                                                                                                                     & 0.53                                                                                                                 & \textbf{0.33}                                            
                                           \\ \hline
\end{tabular}
\end{adjustbox}
\caption{Performance of retrieval models on the test data (English).}
\label{tab:tabel 1}
\end{table}

\begin{table}[]
\begin{adjustbox}{width=\columnwidth,center}
\begin{tabular}{cccc}
\hline
\multicolumn{1}{l}{}                       & \textbf{\begin{tabular}[c]{@{}c@{}} Recall@100\end{tabular}} & \textbf{\begin{tabular}[c]{@{}c@{}} MRR @10 \end{tabular}} & \textbf{\begin{tabular}[c]{@{}c@{}} NDCG@5 \end{tabular}} \\ \hline
\multicolumn{1}{c}{\textbf{SBERT-ID (Saheeh Int.)}}     & 0.32                                                                                                                           & \textbf{0.55}                                                                                                           & \textbf{0.33}                                                                                                       \\
\multicolumn{1}{c}{\textbf{SBERT-ID (Yusuf Ali)}}  & 0.31                                                                                                                           & 0.49                                                                                                           & 0.3                                                                                                       \\

\multicolumn{1}{c}{\textbf{SBERT-ID (al-Hilali)}} & \textbf{0.33}                                                                                                                     & 0.5                                                                                                                 & 0.31                                                                                                       \\
\multicolumn{1}{c}{\textbf{SBERT-ID (Pickthall)}}        &  0.29                                                                                                                     & 0.48                                                                                                                & 0.29                                                                                                     \\
\multicolumn{1}{c}{\textbf{ColBERT-ID (Saheeh Int.)}}       & \textbf{0.33}                                                                                                                    & 0.53                                                                                                                 & \textbf{0.33}                                                             
                                             \\
\multicolumn{1}{c}{\textbf{ColBERT-ID (Yusuf Ali)}}       & 0.28                                                                                                                    & 0.46                                                                                                                 & 0.27  
                                               \\
\multicolumn{1}{c}{\textbf{ColBERT-ID (al-Hilali)}}       & 0.25                                                                                                                     & 0.5                                                                                                                 & 0.3                                 
                                           \\
\multicolumn{1}{c}{\textbf{ColBERT-ID (Pickthall)}}       & {0.27}                                                                                                                     & {0.47}                                                                                                                 & {0.28}                              

                                           \\ \hline
\end{tabular}
\end{adjustbox}
\caption{Comparison of the performance of the retrieval models on the test data for different translations of the Holy Qur'an into English.}
\label{tab:tabel 2}
\end{table}

\section{Preparing a Retrieval Model to Extract Relevant Verses from the Holy Qur'an in Arabic}
\label{sec:Arabic IR}

This section discusses how to address the same problem of designing an efficient neural IR model for extracting relevant verses from the Holy Qur'an in Arabic. Though the goal is essentially the same, the resources to achieve it are quite different in the case of the Arabic Language.
The dataset for testing is the same as the one described in Section ~\ref{sec:test dataset}. We use the queries as they were initially formulated in Arabic by the authors of QRCD \citep{MALHAS2022103068}. For the choice of the metrics, refer to Section ~\ref{sec:metrics}.

\subsection{Choice of Arabic LM to Tackle IR Task in the Islamic Domain}
\label{sec:Arabic LMs}

Due to a lack of manually crafted linguistic resources, Arabic is considered a low- or medium-resource language, depending on the domain of application (\citealp{xue-etal-2021-mt5}; \citealp{abboud-etal-2022-cross}). Recent advances in Arabic NLP have brought a number of LMs pre-trained on Arabic corpora and new datasets translated into Arabic or initially curated in Arabic. Arabic is the language of the Holy Qur'an and the source language of numerous Islamic scholarly works. Moreover, the multi-institutional initiative has offered the Arabic NLP community an Open Islamicate Texts Initiative OpenITI \citep{romanov2019openiti}, an excellent source for pre-training a LM for the Islamic domain. These advantageous conditions for the Islamic domain in Arabic let us skip the preliminary stage of corpus preparation and LM pre-training.

However, there is a benefit in comparing how various Arabic LMs can fit as the backbone of the IR system for the Islamic domain. Table ~\ref{tab:table 3} compares Arabic LMs' efficiency in tackling IR task in the Islamic domain out-of-the-box. We use a sentence-transformers framework to compare LMs to avoid a costly training stage. We add an averaging pooling layer on top of BERT embeddings and convert it into a fixed-sized sentence embedding \citep{reimers-gurevych-2019-sentence}. The same model is utilized to create sentence embeddings for both queries and Qur'anic verses, and then answers to the query are found using the cosine similarity measure. The models are not ready to efficiently handle IR tasks without additional training, yet this approach let us to compare LMs' embeddings out-of-the-box. We include in the comparison the bert-base-uncased model and the BPIT model (evaluation is run on the English translation of QRCD).

As we can see from the table, most of the models perform poorly. We can also observe that pre-training on large amounts of data does not necessarily lead to better performance in IR task. CL-AraBERT performs significantly better than other Arabic LMs, and its performance is similar to the BPIT model. It is plausible that, as in the case of CL-AraBERT \citep{MALHAS2022103068} and the BPIT model, pre-training in a continued approach on a domain-specific corpus with specialized vocabulary starting from the general domain checkpoint helps to tackle IR task in the Islamic domain more efficiently. Another noteworthy observation is that the BPIT model exhibits this performance while pre-trained for a short period and with a small corpus of less than 50M tokens. We assume that contextualized weight distillation might help boost the efficiency during the pre-training stage. The second best performing models are bert-base-uncased and bert-base-arabic. Based on the result of Table ~\ref{tab:table 3}, we choose CL-AraBERT as a backbone model to conduct subsequent experiments with IR task in Islamic Domain in Arabic.

\begin{table}[]
\begin{adjustbox}{width=\columnwidth,center}
\begin{tabular}{cccc}
\hline
\multicolumn{1}{l}{}                       & \textbf{\begin{tabular}[c]{@{}c@{}} Number of tokens/ \\Domain \end{tabular}} & \textbf{\begin{tabular}[c]{@{}c@{}} MRR@10 \end{tabular}} & \textbf{\begin{tabular}[c]{@{}c@{}} NDCG@5 \end{tabular}} \\ \hline
\multicolumn{1}{c}{\textbf{\makecell{bert-base-arabic-\\camelbert-mix \\ \citep{inoue-etal-2021-interplay}}}}     & 17.3B/GD                                                                                                                           & 0.01                                                                                                           & 0.01                                                                                                       \\
\multicolumn{1}{c}{\textbf{\makecell{bert-base-arabic-\\camelbert-ca \\ \citep{inoue-etal-2021-interplay}}}}  & 847M/ID                                                                                                                           & 0.01                                                                                                           & 0.01                                                                                                       \\

\multicolumn{1}{c}{\textbf{\makecell{bert-base-arabertv02\\ \citep{antoun-etal-2020-arabert}}}} & 8.6B/GD                                                                                                                     & 0.01                                                                                                                 & 0.01                                                                                                      \\
\multicolumn{1}{c}{\textbf{\makecell{bert-base-arabic \\ \citep{safaya-etal-2020-kuisail}}}}        &  8.2B/GD                                                                                                                     & 0.06                                                                                                                & 0.02                                                                                                     \\

\multicolumn{1}{c}{\textbf{\makecell{bert-base-uncased \\ \citep{devlin-etal-2019-bert}}}}       & 3.3B/GD                                                                                                                     & 0.07                                                                                                                 & 0.03   
                                                         \\
\multicolumn{1}{c}{\textbf{\makecell{CL-AraBERT \\ \citep{MALHAS2022103068}}}}       & 2.7B+1.05B/GD+ID                                                                                                                     & \textbf{0.11}                                                                                                                 & \textbf{0.06}                                                             
                                    
                                              \\
\multicolumn{1}{c}{\textbf{BPIT}}       & 3.3B+50M/GD+ID                                                                                                                     & \textbf{0.11}                                                                                                               & \textbf{0.06}                                                             

                                           \\ \hline
\end{tabular}
\end{adjustbox}
\caption{Performance of LMs on the test dataset. GD stands for General domain and ID for Islamic domain.}
\label{tab:table 3}
\end{table}

\subsection{Knowledge Distillation Approach to Improve Performance of Arabic LM in IR Task}
\label{sec: Distillation}

The lack of manually crafted linguistic resources in low-resource languages can be tackled by knowledge distillation. \citet{reimers-gurevych-2020-making} showed that it is possible to improve the performance of sentence embedding models by mimicking the performance of a stronger model. They used parallel corpora to teach the student model to produce sentence embeddings close to the embeddings of the teacher model. Their experiment uses the English SBERT model to initialize the teacher model, and multilingual XLM-RoBERTa \citep{conneau-etal-2020-unsupervised} is used as a student model. Our experiment uses the SBERT-BPIT (Section ~\ref{sec:BPIT}) as a teacher model and the bilingual EN-AR student model. The student model combines the embedding matrix of the CL-AraBERT for Arabic tokens and the BPIT model for English tokens, and the encoder weights are borrowed from the BPIT model. We use a combination of parallel datasets (EN-AR) available on the OPUS website \citep{tiedemann-2012-parallel}: TED2020, NewsCommentary, WikiMatrix, Tatoeba, and Tanzil, total size of training data is around 1.1M sentences (for hyperparameters details, see Appendix ~\ref{sec:appendix}). Table ~\ref{tab:table 4} presents the evaluation results of this approach on the test dataset (Bilingual-distilled-EN-AR model). We can see a significant improvement compared to the results of CL-AraBERT from Table ~\ref{tab:table 3}, yet the performance is practically twice lower than the performance of the equivalent English model (SBERT-BPIT, Table ~\ref{tab:tabel 1}). 

\begin{table}[]
\begin{adjustbox}{width=\columnwidth,center}
\begin{tabular}{cccc}
\hline
\multicolumn{1}{l}{}                       & \textbf{\begin{tabular}[c]{@{}c@{}} Recall@100\end{tabular}} & \textbf{\begin{tabular}[c]{@{}c@{}} MRR @10 \end{tabular}} & \textbf{\begin{tabular}[c]{@{}c@{}} NDCG@5 \end{tabular}} \\ \hline
\multicolumn{1}{c}{\textbf{Bilingual-distilled}}     & 0.12                                                                                                                           & 0.26                                                                                                           & 0.15                                                                                                       \\
\multicolumn{1}{c}{\textbf{SBERT-AR-NLI}}  & 0.21                                                                                                                           & 0.38                                                                                                           & 0.21                                                                                                       \\

\multicolumn{1}{c}{\textbf{SBERT-AR-MARCO}} & 0.23                                                                                                                     & 0.4                                                                                                                 & 0.23                                                                                                       \\
\multicolumn{1}{c}{\textbf{ColBERT-AR}}        &  0.28                                                                                                                     & 0.47                                                                                                                & \textbf{0.29}                                                                                                     \\
\multicolumn{1}{c}{\textbf{SBERT-AR-ID}}       & 0.25                                                                                                                     & 0.45                                                                                                                 & 0.27                                                             
                                             \\
\multicolumn{1}{c}{\textbf{ColBERT-AR-ID}}       & \textbf{0.29}                                                                                                                     & \textbf{0.48}                                                                                                                 & \textbf{0.29}                                          
                                           \\ \hline
\end{tabular}
\end{adjustbox}
\caption{Performance of retrieval models on the test dataset (Arabic).}
\label{tab:table 4}
\end{table}

\subsection{ Training on Arabic Natural Language Inference Dataset to Improve Sentence Embeddings}
\label{sec: NLI}
Another approach that can help to improve the quality of the sentence embeddings is training on the Natural Language Inference (NLI) dataset (\citealp {reimers-gurevych-2019-sentence}; \citealp{bowman-etal-2015-large}; \citealp{williams-etal-2018-broad}) . \citet{conneau-etal-2018-xnli} introduced Cross-lingual Natural Language Inference (XNLI) comprising 7500 examples for development and test sets translated into 15 languages, including Arabic. 
We train CL-AraBERT on XNLI following \citet{reimers-gurevych-2019-sentence}, using 400k machine-translated training examples that accompany XNLI development and test set (more details in Appendix ~\ref{sec:appendix}). The performance of this model (SBERT-AR-NLI, Table ~\ref{tab:table 4}) is better than Bilingual-distilled-EN-AR, yet lower than SBERT-BPIT (Table ~\ref{tab:tabel 1}).

\subsection{Employing Machine-Translated Datasets to Overcome The Lack of Large Training Data}
\label{sec:MT}

Although the quality of the machine-translated dataset is inferior to human translation, the accessibility of machine-translated text helps to generate a considerable training set which is essential for preparing a retrieval model. The experiment with training on the XNLI dataset from section ~\ref{sec: NLI} showed that training on a machine-translated dataset can achieve competitive performance. This motivates us to extend this experiment further to the MS MARCO dataset. MS MARCO is a large collection of datasets focused on deep learning in search \citep{bajaj2018ms}, including the IR dataset that comprises more than half a million queries and is accompanied by a collection of 8.8M passages and 39M triplets for training. Another advantage of using MS MARCO, besides a sizable training set, is that it is more suitable for training IR systems, and we can experiment with both SBERT and ColBERT approaches to prepare retrieval models and compare their performance across languages. \citet{bonifacio2022mmarco}
presented a multilingual version of the MS MARCO dataset created using machine translation comprising 13 languages. We use the Arabic translation of MS MARCO and train SBERT-AR-MARCO  equivalently to SBERT-BPIT and ColBERT-AR following the training procedure of ColBERT-BPIT (Section ~\ref{sec:BPIT}). Table ~\ref{tab:table 4} demonstrates that training on MS MARCO can give better results compared to other training approaches described in Sections ~\ref{sec: Distillation} and ~\ref{sec: NLI}.

\subsection{In-domain Training of Retrieval Model for Qur'anic IR in Arabic}

In the last stage, we perform training on in-domain data and repeat the successful experiment with dataset augmentation in English. The steps to augment dataset are the same (see Section ~\ref{SEC:IN-domain}). We use a cross-encoder trained on machine-translated MS MARCO to score ayah pairs, which results in a slightly different count of selected pairs (2723). We continue training SBERT-AR-MARCO and ColBERT-AR on in-domain data and produce SBERT-AR-ID and ColBERT-AR-ID. 

The performance of these retrieval models is included in Table ~\ref{tab:table 4}, and we can observe further improvement after training on in-domain data. The best-performing model is \textbf{ColBERT-AR-ID}, and it is plausible that the retrieval approach of the ColBERT model that leverages more fine-grained interactions between a query and a verse \citep{khattab2020colbert} is especially advantageous for languages with complex morphological structures, such as Arabic. 

\section{Model comparison and Final analysis}
\label{sec:Final}

\begin{figure}[h]
\centering
\includegraphics[width=0.9\columnwidth]{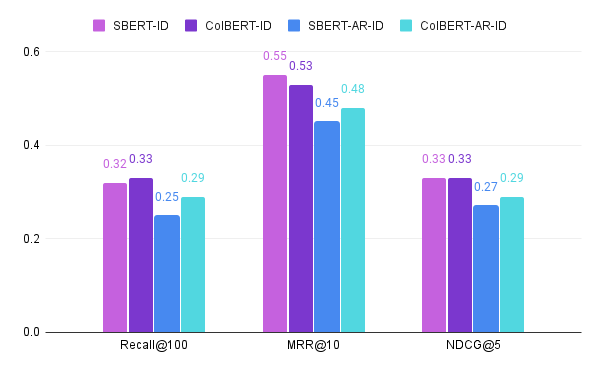}
\caption{Comparison of the retrieval models for the Islamic domain (ID) for English and Arabic across all metrics.}
\label{fig:figure 3}
\end{figure}

\begin{figure}[h]
\centering
\includegraphics[width=0.9\columnwidth]{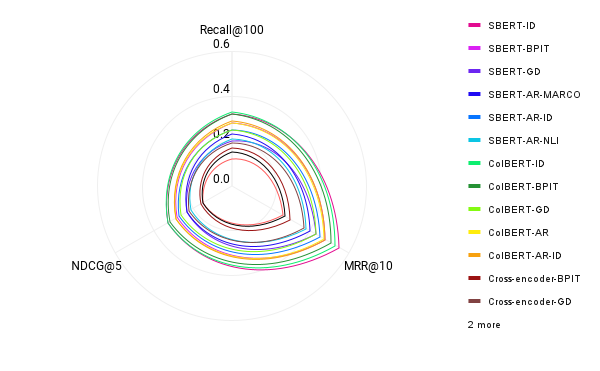}
\caption{Comparison of the retrieval models for the Islamic domain (ID) for English and Arabic across all metrics.}
\label{fig:figure 4}
\end{figure}

\begin{table}[]
\begin{adjustbox}{width=\columnwidth,center}
\begin{tabular}{cccc}
\hline
\multicolumn{1}{l}{}                       & \textbf{\begin{tabular}[c]{@{}c@{}} Recall@100\end{tabular}} & \textbf{\begin{tabular}[c]{@{}c@{}} MRR @10 \end{tabular}} & \textbf{\begin{tabular}[c]{@{}c@{}} NDCG@5 \end{tabular}} \\ \hline
\multicolumn{1}{c}{\textbf{SBERT-AR-ID}}     & 0.25                                                                                                                           & 0.45                                                                                                           & 0.27                                                                                                       \\
\multicolumn{1}{c}{\textbf{ColBERT-AR-ID}}  & 0.29                                                                                                                           & 0.48                                                                                                           & 0.29                                                                                                       \\

\multicolumn{1}{c}{\textbf{SBERT-AR-ID (passages)}} & 0.7                                                                                                                     & 0.47                                                                                                                & 0.35                                                                                                      \\
\multicolumn{1}{c}{\textbf{ColBERT-AR-ID (passages)}}        &  \textbf{0.77}                                                                                                                     & \textbf{0.53}                                                                                                                & \textbf{0.43}                                                                 
                                           \\ \hline
\end{tabular}
\end{adjustbox}
\caption{Performance of Arabic retrieval models on the passage retrieval task (Arabic).}
\label{tab:table 5}
\end{table}

Figure ~\ref{fig:figure 3}  compares all the retrieval models for the Islamic domain (ID) for English and Arabic across all metrics. A noteworthy observation is that all English retrieval models outperform their Arabic equivalents, which can be explained by the complexity of the Arabic language and the usage of machine-translated data. Nevertheless, the results of Arabic retrieval models are not far apart from English models, and specifically, with the employment of the ColBERT model, we can see a competitive performance (\textbf{0.48} for MRR@10 and \textbf{0.29} for Recall@100 and NDCG@5). 

The radar chart (Figure ~\ref{fig:figure 4}) shows a more comprehensive comparison across all models. We can see that the radar chart has a tapered shape overall, with an MRR@10 axis being the most prolonged edge, indicating that all models show the best results for this metric. Moreover, NDCG@5 and Recall@100 are more proportionally placed against each other, signifying that the performance for these metrics is similar across all the models. 
SBERT-ID and ColBERT-ID (magenta and green colors) are located at the edge, showing the best performance. They are followed by ColBERT-BPIT and SBERT-BPIT (English models), and Arabic ColBERT and SBERT models are located in the middle of the chart.
In the center, we can see BM25 and the Bilingual-distilled model, these are models with the lowest performance.

In addition, we conducted tests on two models, ColBERT-AR-ID and SBERT-AR-ID (as shown in Table ~\ref{tab:table 5}), for the passage retrieval task \citep{malhas2023thesis}. We did not apply any passage or query expansion heuristics \citep{malhas2023thesis}. Our findings indicate that this approach is less challenging and increases the MRR@10 score, especially for the ColBERT model. The NDCG@10 score grows by 0.08 for the SBERT model and by 0.14 for the ColBERT model. Moreover, the Recall@100 grows by almost threefold.

\section{Related work}
\label{sec:related work}
\citet{thakur-etal-2021-augmented} proposed a data augmentation technique to train sentence transformers when little data for in-domain training is available. \citet{wang-etal-2021-tsdae-using} and \citet{wang-etal-2022-gpl} experimented with domain adaptation techniques for embedding models.

The topic of the choice of hard negatives is discussed in works of: \citet{qu-etal-2021-rocketqa}, \citet{ren-etal-2021-rocketqav2}, \citet{karpukhin-etal-2020-dense}, \citealp{xiong2021answering}.

\citet{bashir2023arabic} wrote a detailed overview of the state of Qur'anic NLP, including the present state of search and retrieval technologies. Most of the approaches described use keywords-based or ontology-driven search. A few works employ semantic search based on deep-learning methods: \citet{alshammeri2021detecting} use doc2vec; \citet{MOHAMED2022934} utilize word2vec.
\citet{MALHAS2022103068} pre-trained CL-Arabert on OpenITI \citep{romanov2019openiti} starting from the AraBERT checkpoint \citep{antoun-etal-2020-arabert}. They also introduced the first Qur'anic Reading Comprehension Dataset (QRCD) that we used as a test data for the Qur'anic IR task.

\section{Conclusion}
In this paper, we employed state-of-the-art approaches in IR to analyze and compare what works better to improve Qur'anic IR in English and Arabic. The results show that retrieval models in English outperform their Arabic equivalents. The inherent linguistic complexity of the Arabic language may explain this performance gap; nevertheless, transferring successful experiments from English to Arabic, applying large machine-translated datasets, and using the proposed data-augmentation technique helped to enhance the results in Qur'anic IR in Arabic.

One of the possible directions to take in the future is to extend this work to encompass more languages. This would broaden the scope of the semantic search for the Holy Qur'an, making it accessible to a larger audience. Moreover, research conducted in a multilingual environment helps to exchange insights among languages and enhance the results in Qur'anic IR.

Another essential step is to extensively evaluate real-world user queries to analyze models' performance in practice.  \footnote{A live testing system is deployed at \url{rttl.ai}}

\section*{Acknowledgment}
This research would not have been possible without my colleague at rttl.ai, Mohammed Makhlouf, who gathered data for training the language and retrieval models. Additionally, I would like to thank Dr. Rana Malhas and Dr. Tamer Elsayed from Qatar University for helping to verify the quality of the English translation of the queries of the QRCD and anonymous reviewers for their valuable feedback.

\section*{Limitations}
One of the main limitations of our paper is the quality of machine-translated datasets, such as XNLI train set \citep{conneau-etal-2020-unsupervised} and mMARCO \citep{bonifacio2022mmarco} translation into Arabic. Using machine translation is a solution to address a lack of data for training models for low or medium-resource languages like Arabic.
The quality of automated translation is constantly improving and has reached a good quality recently; nevertheless, it is not yet equivalent to the high quality of human translation done by experts in the field. 

\section*{Ethics Statement}
We do not anticipate any considerable risks associated with our work. 
The data and other related resources in this work are publically available, and no private data is involved. We respect previous work done in the field and appropriately cite the methods and datasets we are using. To prevent misuse of pre-trained models, we carefully consider applications and provide access upon request. \footnote{Contact us at \url{ hello@rttl.ai}}

\bibliography{anthology,custom}
\bibliographystyle{acl_natbib}

\appendix

\section{Appendix}
\label{sec:appendix}

\subsection{Hyperparameter details}
\label{sec:hyperparameters}

\begin{table}[h]
\begin{adjustbox}{width=\columnwidth,center}
\begin{tabular}{ccc}
\hline
\textbf{Computing Infrastructure}    & \multicolumn{2}{c}{2 x NVIDIA RTX 3090 GPU}    \\ \hline
\multicolumn{2}{c}{\textbf{Hyperparameter}} & \textbf{Assignment}                     \\ \hline
\multicolumn{2}{c}{number of epochs}        & 10 \\ \hline
\multicolumn{2}{c}{batch size}              & 128                             \\ \hline
\multicolumn{2}{c}{maximum learning rate}   & 0.0005                        \\ \hline
\multicolumn{2}{c}{learning rate optimizer} & Adam                                    \\ \hline
\multicolumn{2}{c}{learning rate scheduler} & None or Warmup linear                   \\ \hline
\multicolumn{2}{c}{Weight decay}            & 0.01                                    \\ \hline
\multicolumn{2}{c}{Warmup proportion}       & 0.06                                    \\ \hline
\multicolumn{2}{c}{learning rate decay}     & linear                                  \\ \hline
\end{tabular}
\end{adjustbox}
\caption{Hyperparameters for continual pre-training of BPIT model.}
\label{tab:appendix-table-a}
\end{table}

    For training SBERT and ColBERT models, we follow training recommendations implemented by the authors. To ensure fair comparison across models and languages, all the hyperparameters for SBERT models are identical, and the same applies to ColBERT models. 
    
\begin{table}[!htbp]
\begin{adjustbox}{width=\columnwidth,center}
\begin{tabular}{ccc}
\hline
\textbf{Computing Infrastructure} & \multicolumn{2}{c}{2 x NVIDIA RTX 3090 GPU} \\ \hline

\multicolumn{2}{c}{\textbf{Hyperparameter}}     & \textbf{Assignment}           \\ \hline
\multicolumn{2}{c}{number of epochs}            & 10                            \\ \hline
\multicolumn{2}{c}{batch size}                  & 64                             \\ \hline
\multicolumn{2}{c}{learning rate}               & 2e-5                          \\ \hline
\multicolumn{2}{c}{pooling}                     & mean                         \\ \hline

\end{tabular}
\end{adjustbox}
\caption{Hyperparameters for training SBERT models.}
\label{tab:appendix-table-b}
\end{table}

\begin{table}[!htbp]
\begin{adjustbox}{width=\columnwidth,center}
\begin{tabular}{ccc}
\hline
\textbf{Computing Infrastructure} & \multicolumn{2}{c}{2 x NVIDIA RTX 3090 GPU} \\ \hline

\multicolumn{2}{c}{\textbf{Hyperparameter}}     & \textbf{Assignment}           \\ \hline
\multicolumn{2}{c}{number of epochs}            & 1                             \\ \hline
\multicolumn{2}{c}{batch size}                  & 32                             \\ \hline
\multicolumn{2}{c}{learning rate}               & 1e-5                          \\ \hline
\end{tabular}
\end{adjustbox}
\caption{Hyperparameters for training ColBERT models.}
\label{tab:appendix-table-b}
\end{table}

\begin{table}[!htbp]
\begin{adjustbox}{width=\columnwidth,center}
\begin{tabular}{ccc}
\hline
\textbf{Computing Infrastructure} & \multicolumn{2}{c}{2 x NVIDIA RTX 3090 GPU} \\ \hline

\multicolumn{2}{c}{\textbf{Hyperparameter}}     & \textbf{Assignment}           \\ \hline
\multicolumn{2}{c}{number of epochs}            & 1                             \\ \hline
\multicolumn{2}{c}{batch size}                  & 32                             \\ \hline
\multicolumn{2}{c}{learning rate}               & 2e-5                          \\ \hline

\end{tabular}
\end{adjustbox}
\caption{Hyperparameters for training Cross-encoders.}
\label{tab:appendix-table-b}
\end{table}

\end{document}